\documentclass[3p]{elsarticle}

\usepackage{amssymb}

\usepackage{graphicx}
\usepackage{balance}
\usepackage{color}
\usepackage{xcolor}
\usepackage{multirow}
\usepackage{caption}

\usepackage{subfigure,epsfig,amsfonts,amsmath,cases,amssymb,graphics, latexsym, times, algorithmic, algorithm, bbm, color, soul,wrapfig,comment}

\journal{Smart Health}

\begin{document}

\begin{frontmatter}



\title{Early hospital mortality prediction using vital signals}


\author[wm]{Reza Sadeghi\corref{cor1}}
\ead{sadeghi.2@wright.edu}

\author[wm]{Tanvi Banerjee}
\ead{tanvi.banerjee@wright.edu}

\author[mines]{William Romine}
\ead{william.romine@wright.edu }

\address[wm]{Department of Computer Science and Engineering, Kno.e.sis Research Center, Wright State University, Dayton, OH, USA}

\address[mines]{Department of Biological Sciences, Wright State University, Dayton, OH, USA}

\cortext[cor1]{Corresponding author}

\begin{abstract}
Early hospital mortality prediction is critical as intensivists strive to make efficient medical decisions about the severely ill patients staying in intensive care units (ICUs). As a result, various methods have been developed to address this problem based on clinical records. However, some of the laboratory test results are time-consuming and need to be processed. In this paper, we propose a novel method to predict mortality using features extracted from the heart signals of patients within the first hour of ICU admission. In order to predict the risk, quantitative features have been computed based on the heart rate signals of ICU patients suffering cardiovascular diseases. Each signal is described in terms of $12$ statistical and signal-based features. The extracted features are fed into eight classifiers: decision tree, linear discriminant, logistic regression, support vector machine (SVM), random forest, boosted trees, Gaussian SVM, and K-nearest neighborhood (K-NN). To derive insight into the performance of the proposed method, several experiments have been conducted using the well-known clinical dataset named Medical Information Mart for Intensive Care III (MIMIC-III). The experimental results demonstrate the capability of the proposed method in terms of precision, recall, F1-score, and area under the receiver operating characteristic curve (AUC). The decision tree classifier satisfies both accuracy and interpretability better than the other classifiers, producing an F1-score and AUC equal to $0.91$ and $0.93$, respectively. It indicates that heart rate signals can be used for predicting mortality in patients in the care units especially coronary care units (CCUs), achieving a comparable performance with existing predictions that rely on high dimensional features from clinical records which need to be processed and may contain missing information.

\end{abstract}

\begin{keyword}
intensive care, mortality prediction, statistical and signal-based features
\end{keyword}

\end{frontmatter}

\section{Introduction}
  \label{sec:intro}

Intensive care unit (ICU) is a ward in hospital, where seriously ill patients are cared for by specially trained staff.  Quick and accurate decisions for the patients are needed. As a result, a wide range of decision support systems have been deployed to aid intensivists for prioritizing the patients who have a high risk of mortality.

Most mortality prediction systems are considered as score-based models~\cite{calvert_using_2016}\cite{simpson_new_2016}\cite{le_gall_new_1993}\cite{knaus_apache_1985} which appraise disease severity to predict an outcome. These models utilize patient demographics and physiological variables such as age, temperature, and heart rate collected within the initial $12$ to $24$ hours after ICU admission with the aim of assessing ICU performance. The score-based models employ certain features that sometimes are not available at ICU admission. Also, they make decisions according to a collection of data after at least first $12$ hours of ICU admission. To enhance the proficiency, the customized models refine the score-based models for usage within specific conditions. For instance,~\cite{dervishi_fuzzy_2017} introduces a model to predict the risk of mortality due to cardiorespiratory arrest. Although these models provide adequate results, the ICU patients are varied and subjected to multiple diseases. Therefore, selecting the right model for a special patient who is immediately admitted to ICU is difficult. On the other hand, various studies~\cite{awad_early_2017}\cite{wojtusiak_c-lace:_2017}\cite{ribas_severe_2011}\cite{kim_comparison_2011}\cite{purushotham_benchmark_2017} express the superiority of data mining techniques over traditional score-based models. The data mining models have exerted different techniques such as  random forest~\cite{awad_early_2017}\cite{wojtusiak_c-lace:_2017}, support vector machine~\cite{ribas_severe_2011}, decision tree~\cite{kim_comparison_2011}, and deep learning~\cite{purushotham_benchmark_2017}\cite{avati_improving_2017}\cite{beaulieu-jones_mapping_2017}\cite{song_attend_2017}. Furthermore, some of the methods like \cite{venugopalan_combination_2017} engage a pipeline of data mining techniques to predict the risk of mortality. These methods are organized based on certain clinical records which are collected in initial hours after ICU admission. However, laboratory test results need to be processed and many clinical records contain missing values~\cite{yadav_mining_2018}. While vital signals can provide numerous information which has been proven to possess strong relation with the mortality~\cite{zhang_resting_2015}. Therefore, vital signal fluctuations can provide high capability to predict the mortality risk more accurately and faster than clinical-based methods.

The main goal of this paper is to provide an early mortality prediction of patients based on their first hour after ICU admission according to their heart rate signals. Our study relies on the Medical Information Mart for Intensive Care III, MIMIC-III Waveform Database records~\cite{johnson_mimic-iii_2016}. We propose a method to extract both statistical and signal-based features from the heart signals and employ well-known classifiers such as logistic regression and decision tree to predict hospital mortality, i.e. death inside the hospital.

The rest of the paper is organized as follows: Section~\ref{sec:related_work} presents a literature review on the related studies. Section~\ref{sec:methodology} describes the proposed method in four subsections of data description, signal preprocessing, feature extraction, and classification. To evaluate the performance of the proposed method, Section~\ref{sec:experiments} is allocated to the experiments and discussions. Finally, Section~\ref{sec:conclusion} summarizes the conclusion and future work.

\section{Related Work}
  \label{sec:related_work}

There is an increasing interest in addressing early hospital mortality prediction. The proposed systems can be categorized into three classes of score-based, customized, and data mining models.

Various score-based approaches such as acute physiology and chronic health evaluation (APACHE)~\cite{knaus_apache_1985}, simplified acute physiology score (SAPS)~\cite{le_gall_new_1993}, and quick sepsis-related organ failure assessment score (qSOFA)~\cite{simpson_new_2016} have been proposed. APACHE score is the best-known and widely used in intensive cares~\cite{vincent_critical_2010}. The original APACHE score~\cite{knaus_apache-acute_1981} employed 34 physiological measures from initial 24 hours after ICU admission to determine the chronic health status of the patients.~\cite{knaus_apache_1985} introduced the APACHE II scoring model including a reduction in the number of variables to $12$ routine physiological measurements, along with the age of patients. Extending that, the APACHE III improved the effectiveness of mortality prediction by adding new variables such as race, length of stay in ICU, and prior place before ICU. APACHE IV also endeavored to enhance the over prediction problem of the APACHE III by adding new variables and using the weights utilized in APACHE III~\cite{zimmerman_acute_2006}. The traditional severity of illness score-based models commonly attempted to predict based on either specific age ranges, or information recorded within the first $24$ hours of ICU admission~\cite{johnson_reproducibility_2017}. Furthermore, they utilized features which are not always available at the time of ICU admission. For instance, the APACHE IV applied its analysis on over $100$ variables like chronic health variables of AIDS, cirrhosis, hepatic failure, immunosuppression which may not be recorded at the time of admission.

The customized models make a decision according to the characteristics of either specific health problems such as cardiorespiratory arrest~\cite{dervishi_fuzzy_2017} and early severe sepsis~\cite{le_gall_customized_1995}, or specific geographical areas such as France~\cite{le_gall_mortality_2005} or Australia~\cite{metnitz_austrian_2009}. For instance, Le Gall and coworkers~\cite{le_gall_mortality_2005} customized the SAPS II model based on the French patients' characteristics. They used the logit of the original SAPS II model and computed the coefficients according to the data. Furthermore, they tried to expand the second version of SAPS by adding six variables (age, sex, length of hospital stay before ICU admission, and the patient's location before ICU) that are potentially associated with mortality. Although these models provide adequate results, most ICU patients are elderly people over $65$ years~\cite{banerjee_validating_2017} who are faced with multiple ailments. Also, selecting the right model is challenging due to the variety of patients who are immediately admitted to ICU. Moreover, the models for specific geographical areas are not extendable for other cases. 

The third class of methods employ data mining techniques to forecast mortality. For instance,~\cite{awad_early_2017} devised a method based on random forest and the synthetic minority over-sampling technique. In another method, Venugopalan et. al~\cite{venugopalan_combination_2017} used a pipeline of logistic regression, neural network, and conditional random forest. The three categories of demographic, lab, and chart data such as gender, age, height, sodium, creatinine, and heart rate have been fed to logistic regression, neural network, and conditional random forest, respectively. These methods focus on using clinical records instead of waveform data while in practice, many clinical records such as laboratory test results need to be processed which could delay the clinical decision support process. 

To address these issues, we propose a method for early mortality prediction of patients based on the first hour after ICU admission using heart rate signals. To the best of our knowledge, this paper is the first work which utilizes only heart signals for early mortality prediction using the MIMIC-III dataset. We describe each signal in terms of $12$ statistical and signal-based features which are fed into multiple transparent and non-transparent classifiers.

\section{Methodology}
  \label{sec:methodology}

This section presents a novel method which utilizes statistical and signal-based features with the purpose of fast and accurate early hospital mortality prediction. Subsection~\ref{sub:data_description} provides a review on the MIMIC-III clinical dataset while subsections~\ref{sub:signal} and~\ref{sub:feature} describe signal preprocessing and feature extraction, respectively. Ultimately, subsection~\ref{sub:classification} presents an overview on the descriptive classifiers employed to predict whether a patient survives or passes away based on the characteristics of their ECG signal.

\subsection{Data Description}
  \label{sub:data_description}

\begin{wrapfigure}{R}{0.46\textwidth}
\centering
\includegraphics[width=0.46\textwidth]{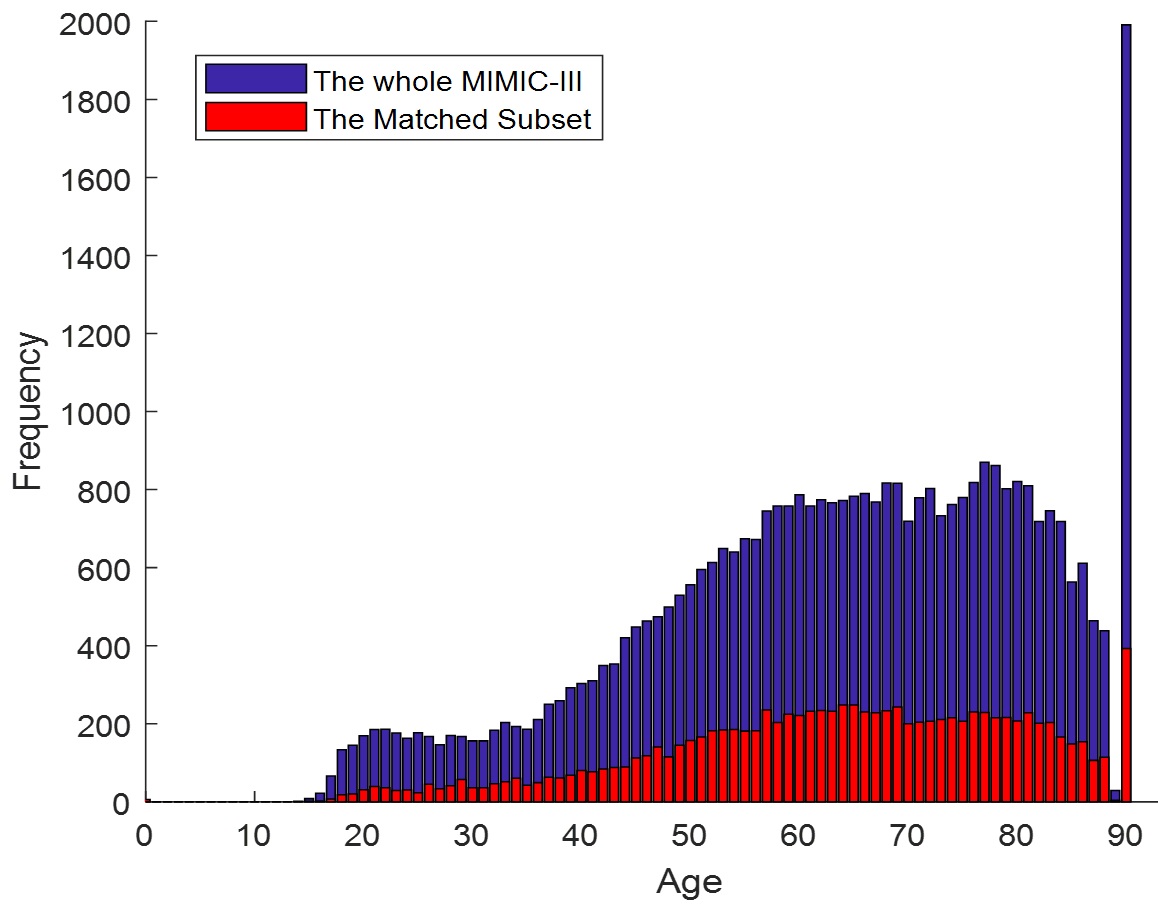}
\caption{The age distribution over the Whole MIMIC-III (without infants) and the Matched Subset}
\label{fig:Age_distribution}
\end{wrapfigure}

This study is conducted over the well-known MIMIC-III database comprising the records of $46520$ patients who stayed in critical care units. Due to the de-identification process, there are only $10282$ patients whose the clinical data in the MIMIC-III are associated with the related vital signals in the Matched Subset. As shown in the Figure~\ref{fig:Age_distribution}, the age distributions of the whole MIMIC-III (without infants) and the Matched Subset are similar. Hence, the outcomes of the Matched Subset can be extended to the whole database. It is worth mentioning that due to the de-identification process, all the patients greater than or equal to $90$ years of age are assigned to one group.

Also, the hospital wards for patients throughout their hospital stay have been reported via the transfers table in the clinical dataset. Indeed, it specifies which of the care units described in Table~\ref{table:care_units} have been allocated to each patient in a certain time. Since nearly $90$ percent of patients in the Matched Subset suffer from cardiovascular diseases, we have focused on predicting the risk of mortality among patients who stayed in coronary care unit (CCU) in this study. CCU is an ICU that takes patients with cardiac conditions required continuous monitoring and treatment.

\begin{table}[H]
\caption{Care Units in MIMIC-III}
\centering
\begin{tabular}{cccccc}
\hline
   \textbf{Care unit}  & \textbf{Description}\\
   \hline
   \vspace{5pt} 
   CCU  & Coronary care unit\\
   \vspace{5pt} 
   CSRU & Cardiac surgery recovery unit\\
   \vspace{5pt}    
   MICU & Medical intensive care unit\\
   \vspace{5pt} 
   NICU & Neonatal intensive care unit\\
   \vspace{5pt} 
   NWARD & Neonatal ward\\
   \vspace{5pt} 
   SICU   & Surgical intensive care unit\\    
   \multirow{1}{*}{TSICU}   &  Trauma/surgical intensive care unit\\
   \hline
\end{tabular}
\label{table:care_units}
\end{table}

\subsection{Signal Preprocessing}
  \label{sub:signal}

The recorded physiological signals are always accompanied with noise due to different recording systems. The MIMIC-III database is extracted from the CareVue and MetaVision clinical information systems provided by Philips and iMDSoft, respectively~\cite{johnson_mimic-iii_2016}. After extracting the data, we truncated the tails which contain only zeros or undefined values. Following this, we replaced the missing values with the previous known ones. Finally, the smoothed version of heart rate signal, $S'(t)$, was computed according to the moving average filter with one-hour windows size $\rho$ in the form of Equation~\ref{equ:smoothing}.  

\begin{equation}
\label{equ:smoothing}
S'(t) = \begin{cases}
  \vspace{5pt} 
  \dfrac{1}{T} $$\sum_{t=1}^{T} S(t) & \rho>=T>=1 \\
  \vspace{5pt} 
  \dfrac{1}{\rho}$$\sum_{t=T}^{T-\rho+1} S(t) & L-\rho>=T>=\rho \\
  \vspace{5pt} 
  \dfrac{1}{T}$$\sum_{t=L-\rho+1}^{L} S(t) & L-\rho+1>=T>=L \\
\end{cases}
\end{equation}

\begin{wrapfigure}{R}{0.46\textwidth}
\centering
\includegraphics[width=0.46\textwidth]{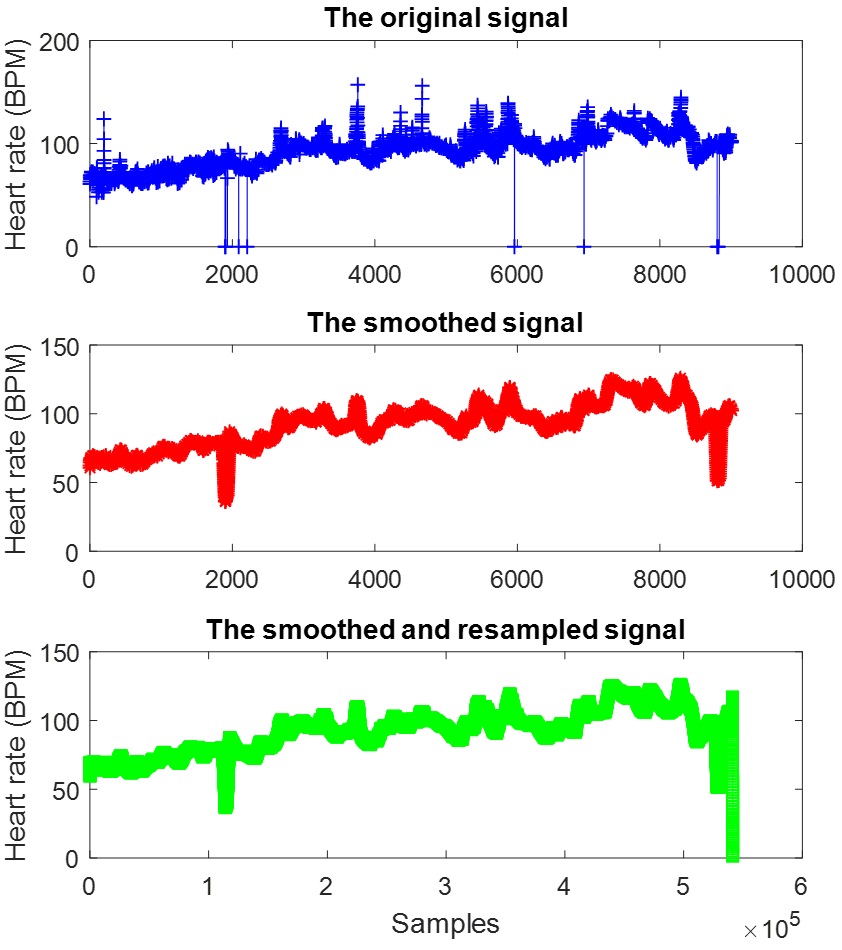}
\caption{The preprocessed heart rate signal of one survived patient from CCU}
\label{fig:Smoothing_signals}
\end{wrapfigure}

where the original signal $S(t)$ contains $L$ samples. On the other hand, the heart signals were recorded with different lengths and sampling rates. For instance, the sampling rate of the heart rate (HR) signals are varied from $1$ to $0.17$ Hz in MIMIC-III database. To avoid biased comparison among signals due to the different sampling rates and lengths, the anti-aliasing finite impulse response (FIR) low-pass filter~\cite{hentschel_digital_1999} was performed over the low sampling rate signals. Indeed, a linear-phase FIR filter interpolates new samples to resample the signals with a lower sampling rate. For instance, as shown in Figure~\ref{fig:Smoothing_signals} the noise samples have been removed by applying the moving average over the original signal. Then, the oversampling method increases the frequency of the heart rate signal to $1 Hz$, leading to increasing the number of samples from $9021$ to $5413×10^5$.

\subsection{Feature Extraction}
  \label{sub:feature}

In order to predict the risk of mortality after the first hour of ICU admission, quantitative features have been computed based on the HR signals. Each signal is described in terms of $12$ statistical and signal-based features which were extracted from the patient's ECG signal. The statistical features reveal useful information about the distributions of the processed data described in the subsection~\ref{sub:signal}. Signal preprocessing. Maximum, minimum, and range can demonstrate the spectrum in which the distribution lies. The skewness indicates whether the distribution is symmetric or skewed.  The kurtosis measures the thickness of the tails of the distribution and the standard deviation shows how the data samples scatter around the mean. Table~\ref{table:features} indicates the average of each feature for both passed away and living patients. The reported values indicate the capability of these features in segregating the two groups of patients based on the proposed statistical and signal-based features. 

\begin{table}[H]
\caption{Descriptive Statistics for Statistical and Signal-Based Features}
\centering
\begin{tabular}{cccccc}
\hline
   \textbf{Column}  & \textbf{Feature} & \textbf{Passed away patients} & \textbf{Alive patients}\\
   \hline
   \vspace{5pt} 
   1 & Maximum & 97.82  & 90.92\\
   \vspace{5pt} 
   2 & Minimum & 80.69  & 76.24\\
   \vspace{5pt}    
   3 & Mean & 88.46  & 81.92\\
   \vspace{5pt} 
   4 & Median & 88.45  & 81.81\\
   \vspace{5pt} 
   5 & Mode & 85.25  & 79.98\\
   \vspace{5pt} 
   6 & Standard deviation & 2.63  & 2.25\\
   \vspace{5pt} 
   7 & Variance & 15.84  & 11.56\\
   \vspace{5pt} 
   8 & Range & 17.13  & 14.68\\
   \vspace{5pt} 
   9 & Kurtosis &17.48  & 17.85\\
   \vspace{5pt} 
   10 & Skewness & 0.83  & 1.02\\
   \vspace{5pt} 
   11 & Averaged power & 8186.02  & 7045.04\\
   \multirow{1}{*}{12}   &  Energy spectral density & 5114.78  & 4420.38\\
   \hline
\end{tabular}
\label{table:features}
\end{table}

The signal-based features in this study fall into two different groups of averaged power and power spectral density~\cite{cohen_autonomic_2000}. The averaged power of a finite discrete-time signal is defined as the mean of the signal's energy. The averaged power of a discrete-time signal $S[n]$ is computed as:

\begin{equation}
\label{equ:average_power}
\bar{P} = \dfrac{E}{n_2 - n_1 +1} = \dfrac{1}{n_2 - n_1 +1} \sum_{n_1}^{n_2} S[n]^2
\end{equation}

where $n_1$ and $n_2$ are the first and last samples, respectively. The signal power is computed by taking the integral of the power spectral density (PSD) of a signal over the entire frequency space. The PSD is the Fourier transform of the biased estimate of the autocorrelation sequence. The PSD of the signal $S[n]$ with sampling rate $\rho$, in the interval $\Delta T$ can be computed as follows:

\begin{equation}
\label{equ:psd}
\bar{P} = \dfrac{\Delta T}{N} \mid \sum_{n=0}^{N-1} S[n]e^{-i2\pi\rho} \mid
\end{equation}

\subsection{Classification}
  \label{sub:classification}

In the MIMIC III dataset, the number of patients who passed away inside the hospital is relatively small in comparison with the number of patients who survived, meaning the dataset is imbalanced. The ratio of physiological signals pointing to the passed away patient in contrast to those who survive is equal to $7.03$. Thus, the early mortality prediction systems are faced with an imbalanced dataset. To handle this issue, a wide range of techniques such as resampling~\cite{awad_early_2017}, cost sensitive classifiers~\cite{perry_imbalanced_2015}, and one-class classifiers~\cite{sadeghi_automatic_2018}\cite{hamidzadeh_weighted_2017} have been proposed. Resampling methods make no assumptions about the distribution of samples and therefore, they can be applicable to any classification problem. Also, they are less sensitive to outliers than other techniques. In this study, we utilize a resampling method called adaptive semi-unsupervised weighted oversampling (A-SUWO)~\cite{nekooeimehr_adaptive_2016} to balance the dataset. 

The $10$-fold cross-validation strategy was used to evaluate the performance of classifiers on the same dataset. In this way, samples are arbitrarily divided into ten disjoint sections. In ten iterations, nine folds shape a group of samples used to train classifiers. Furthermore, the remaining one is utilized to test the learning process. The mean of learning rates determines the performance of the methods in segregation of classes.

In this study, two categories of classifiers are examined: transparent or interpretable models, and non-transparent or black-box models. Transparent classifiers such as decision tree, linear discriminant, logistic regression, and support vector machine (SVM) using the linear kernel explain hidden clinical implications and integrate background knowledge into analysis. Also, they are not only easy to interpret and fast, but also need small memory in practice. On the other hand, non-transparent classifiers like random forest, K-NN, boosted tree, and Gaussian SVM are black-box methods which frequently provide adequate classification results. However, these non-transparent classifiers suffer from lack of easily-comprehensible descriptions for the relations between input and output variables.

\section{Experiments and Results }
  \label{sec:experiments}

In these experiments, a retrospective analysis on patients who stayed in CCU was performed using the information recorded in from the MIMIC-III Waveform Database Matched Subset. This dataset contains the records of $365$ patients who passed away while staying at CCU and $2614$ patients successfully discharged. As mentioned above, the effect of noise samples was reduced by smoothing the heart rate signals using the averaged smoothing filter. Also, resampling of low-sampled signals was used to have a fair comparison. Eventually, the combination of statistical and signal-based features after normalization was fed to several interpretable and non-transparent classifiers which are easy to interpret and statistically powerful, respectively.
 
Four transparent classifiers: decision tree, linear discriminant, logistic regression, and support vector machine (SVM) were examined. The decision tree was implemented based on a CART tree algorithm~\cite{breiman_classification_1984} with Gini's diversity index ($GDI$) as a split criterion. This splitting criterion is one of the most popular impurity measurements which not only performs similar to information gain in most cases~\cite{raileanu_theoretical_2004}, but also has lower computational complexity as a result of avoiding use of the logarithm. The Gini index in the form of Equation~\ref{equ:gdi} is utilized to select the next feature at each node of the tree for splitting the data.

\begin{equation}
\label{equ:gdi}
GDI = 1-\sum_{i} (p(i))^{2}
\end{equation}

where $p(i)$ is the observed fraction of samples in the node, which are labeled as $i$. Therefore, the $GDI$ equal to zero points out to a pure node which contains samples of one class. On the other hand, the $GDI$ for binary classification is equal to $0.5$ at most when a node contains samples of both classes with identical numbers. Furthermore, the linear SVM working based on dot product kernel is a simple linear classifier. As a result, this version of SVM is both easy to be interpreted and fast in prediction.

Regarding to the non-transparent classifiers, four black-box methods of random forest, boosted trees, Gaussian SVM, and K-nearest neighborhood (K-NN) are employed. The random forest and boosted trees utilize $60$ decision tree learners according to the bootstrap aggregating~\cite{breiman_random_2001} and adaptive boosting~\cite{margineantu_pruning_1997} ensemble methods, respectively. Moreover, the Gaussian SVM uses radial basis function kernel and K-NN exerts the K equal to $100$. All the experiments are implemented in MATLAB $9.2.0.538062 (R2017a)$ on the same machine with an Intel processor 2.50 GHz with 8 GB RAM

\subsection{Results}
  \label{sub:results}

The outputs of classifiers can be summarized in four groups: the patients who are truly diagnosed as passed away (TP), the people who are incorrectly labeled as passed away (FP), the records correctly detected as information belonging to survived patients (TN), and finally the ones incorrectly assigned as living patients (FN). These four groups can be aggregated in different ways. 

Equation~\ref{equ:pre} indicates the precision metric as the fraction of patients who have been truly diagnosed as passed away over all the patients predicted as having passed away.  Indeed, the larger number of patients incorrectly predicted as passed away leads to the lower precision for the classifier. Moreover, to see the ability of the classification method in predicting all passed-away patients, we utilize the recall metric presented in Equation~\ref{equ:rec}. In other words, this metric presents the fraction of the patients who are correctly predicted as passed-away over the whole number of passed-away patients.

\begin{equation}
\label{equ:pre}
Precision = \dfrac {TP}{TP + FP}
\end{equation}

\begin{equation}
\label{equ:rec}
Recall = \dfrac {TP}{TP + FN}
\end{equation}

It is worth mentioning that all samples being assigned to positive group lead to high recall and low precision. Then, the harmonic average of precision and recall called F1-score is also considered. Indeed, F1-score described in Equation~\ref{equ:f1} calculates the quality of classification for both passed away and living patients, simultaneously.

\begin{equation}
\label{equ:f1}
F1-score = \dfrac {2\times (Precision\times Recall)}{Precision + Recall}
\end{equation}

\begin{table}[H]
\caption{Classification Results for CCU Mortality}
\centering
\begin{tabular}{cccccc}
\hline
\textbf{Classifier}  & \textbf{Precision} & \textbf{Recall} & \textbf{F1-score} & \textbf{Interpretability}\\
\hline
\vspace{5pt}
\textbf{Random forest} & \textbf{0.97} & \textbf{0.97} & \textbf{0.97} & \textbf{Hard}\\
\vspace{5pt}
Gaussian SVM & 0.95 & 0.96 & 0.96 & Hard\\
\vspace{5pt}
\textbf{Decision tree} & \textbf{0.90} & \textbf{0.92} & \textbf{0.91} & \textbf{Easy}\\
\vspace{5pt}
Boosted trees & 0.91 & 0.83 & 0.87 & Hard\\
\vspace{5pt}
K-NN & 0.80 & 0.85 & 0.82 & Hard\\
\vspace{5pt}
Logistic regression & 0.77 & 0.67 & 0.72 & Easy\\
\vspace{5pt}
Linear discriminant & 0.78 & 0.66 & 0.71 & Easy\\
\multirow{1}{*}{Linear SVM }& 0.80 & 0.63 & 0.70 & Easy\\
   \hline
\end{tabular}
\label{table:results}
\end{table}

As shown in the Table~\ref{table:results}, the decision tree outperforms all transparent classifiers which are easily interpretable and provide some clinical insights into the classification process. Also, the values for F1-score among the transparent classifiers demonstrate a big gap between the decision tree and the others. The F1-score of linear discriminant, linear SVM, and logistic regression is near to $0.71$ while the decision tree results in $0.91$. The linear discriminant assumes that different groups of data are generated based on different Gaussian distributions. However, the amounts of Skewness and Kurtosis of both passed away and surviving patients are not equal to zero (table~\ref{table:features}) which indicates non-Gaussian distribution for the both groups of patients. This is the likely reason why the linear discriminant results in low performance. In addition, weak performance of the logistic regression and linear SVM may indicate that the data are not linearly separable. Furthermore, the performance of these supervised methods is similar to the results of the other empirical comparisons such as~\cite{caruana_empirical_2006} describing that random forest can outperform other classifiers like SVM and K-NN in certain conditions.

\begin{wrapfigure}{R}{0.46\textwidth}
\centering
\includegraphics[width=0.46\textwidth]{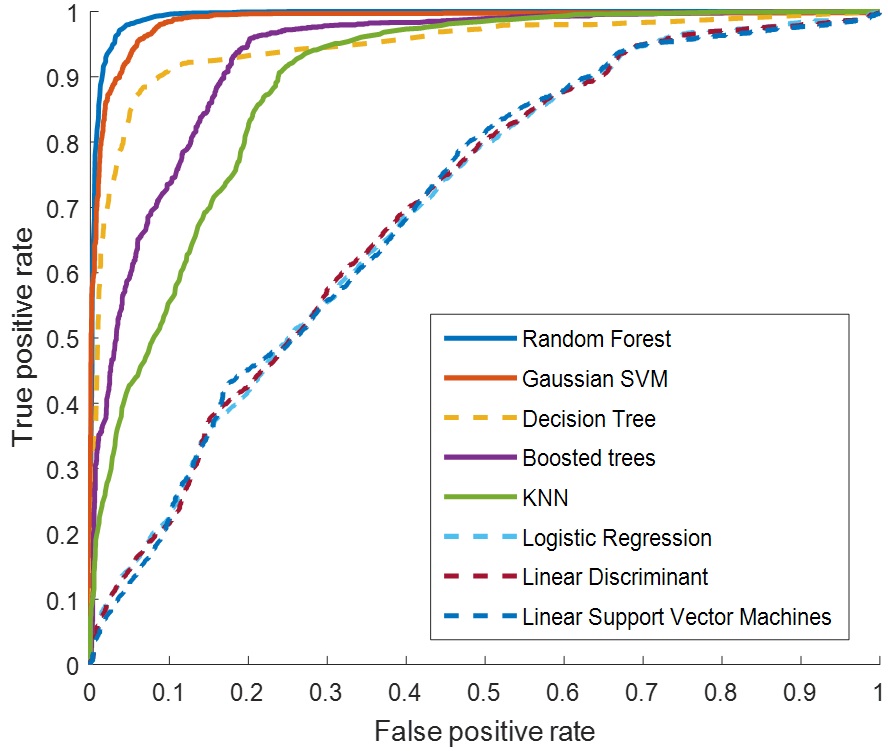}
\caption{The ROC curves of transparent and black-box classifiers shown by dashed and solid lines, respectively}
\label{fig:ROC_curve}
\end{wrapfigure}

From another point of view, all interpretable classifiers except the decision tree have lower recall (near $0.65$) rather than their precision. However, the decision tree has both high precision and recall that shows not only most of the passed-away patients have been correctly recognized but also most of the predicted passed-away patients are correctly assigned to the correct category. As expected, most of the non-transparent classifiers achieve higher performance in comparison to the interpretable classifiers. In addition, random forest comprising several decision tree learners performs better than the other black boxe methods. The interesting point is that the decision tree exceeds many of the non-transparent classifiers including K-NN and boosted tree.

Decision support systems are required to be accurate and robust; however, they also should be interpretable, transparent, and capable of integrating clinical background knowledge into the analysis. Hence, we focus on transparent classifiers and scrutinize their performance in different thresholds. Figure~\ref{fig:ROC_curve} demonstrates that the decision tree outperforms the other transparent classifiers in terms of AUC. Furthermore, the linear SVM, logistic regression, and linear discriminant have similar performance even on different thresholds, which lie lower than the AUC of the decision tree.

Referring to the ROC curve of black-box methods plotted by solid lines in Figure~\ref{fig:ROC_curve}, random forest has the best performance in comparison to Gaussian SVM, boosted trees, and K-NN. Moreover, the curves indicate that random forest and Gaussian SVM have a homogeneous ratio of true positive rate over false positive rate. Furthermore, the ROC curve of decision tree represents the outperformance of this transparent classifier over two black-box methods of K-NN and boosted trees.

The results reveal that the most non-transparent classifiers achieve higher discrimination power while they failed to provide adequate explanations about how the classification results are derived. On the other hand, the interpretable classifiers often attempt to create a decision boundary using the value of linear combination of the sample features. However, most real samples originate from a complex system such as human body. Hence, the decision tree may provide the best choice as a tradeoff between transparency and accuracy. The decision tree discovers knowledge which can be expressed in a readable form while its classification performance is comparable with other methods, even popular non-transparent classifiers.

\subsection{Discussion}
  \label{sub:discussion}

\begin{wrapfigure}{R}{0.46\textwidth}
\centering
\includegraphics[width=0.46\textwidth]{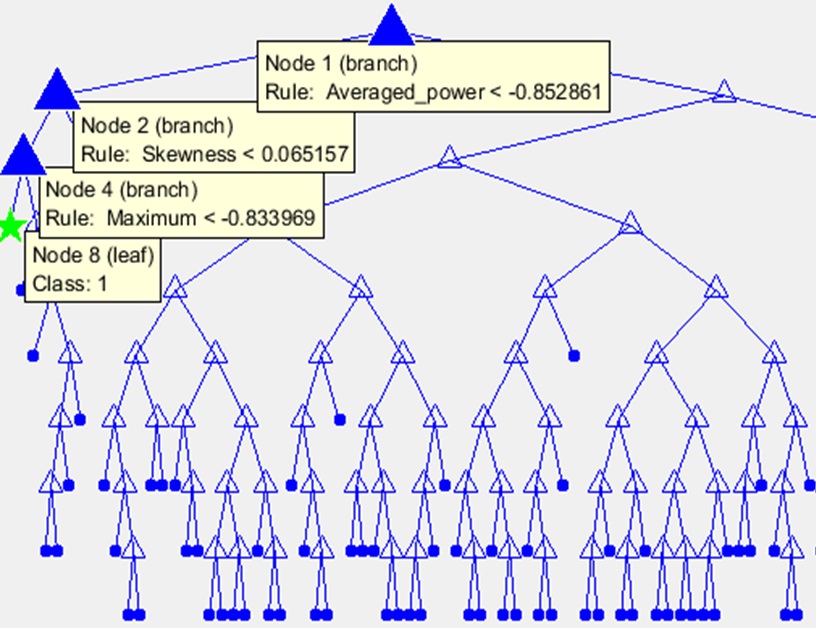}
\caption{The decision tree model comprises decision and leaf nodes represented by triangles and the dots, respectively. The green star shows the left-most leaf node of the model.}
\label{fig:DT}
\end{wrapfigure}

In order to interpret the decision tree qualitatively, Figure~\ref{fig:DT} illustrates the best trained structural model of this classifier gained in the experiments. The tree model hierarchically separates data according to the features leading to a more stable and pure tree. For instance, the left-most child of the decision tree displayed by green star contains records from class $1$ (survived patients). The highlighted path shows records which satisfy the three rules shown in the graph. The first rule divides samples according to the amount of energy spectral density computed for each record. The samples with energy spectral density lower than $-0.85$ are passed to the decision Node $2$ which provides a rule for the amount of Skewness of signals. Node $4$ then filters the samples with value of Maximum less than $-0.83$ which will be assigned to the green star node.

The averaged power, the 11th feature, has been selected as the root of the tree with the highest Gini index. This result shows that using the averaged power features may be promising in early mortality predictions. To further scrutinize the effects of individual features in the decision tree, the estimate of predictor importance is computed. It sums up changes in the risk caused by splits on every independent variable and divides the total result by the number of branch nodes (the tree nodes without any children). Indeed, this sum is taken over the best splits found at each branch node. The importance of features according to this separation is computed as the difference between the risk for the parent node and the sum of risks for its children.

The risk of splitting for each node is composed of the impurity measurement and the node probability. As explained before, we employed the Gini index as the impurity measurement which has less computational complexity in comparison to the information gain. Also, node probability is defined as the number of records reaching the node, divided by the total number of records. Then, the risk of splitting for node x is computed as follows.

\begin{equation}
\label{equ:risk}
Risk(x) = GDI(x) \times Probability(x)
\end{equation}

The estimate of predictor importance for a certain feature is directly associated with the $GDI$ gap between the node corresponding to that feature and its children. This estimation assigns higher importance to features which lead to the largest number of pure children (i.e. terminal nodes). This estimation allots greater importance to the features which have influence on a larger portion of the records. As a result, the feature comprising the root node (in this case the Averaged Power from Figure~\ref{fig:DT}) has higher probability than other features that define rules at lower levels. It allows the feature of the root node to be considered as one of the most important features.

\begin{wrapfigure}{R}{0.46\textwidth}
\centering
\includegraphics[width=0.46\textwidth]{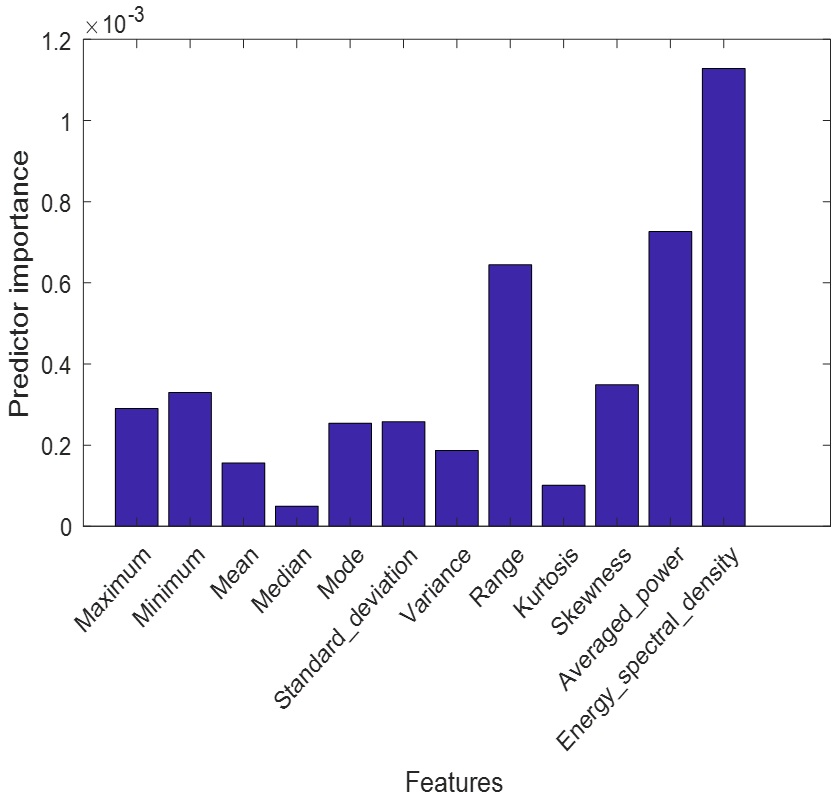}
\caption{Feature importance in the proposed model for mortality prediction based on heart rate signal}
\label{fig:Predictor_importance}
\end{wrapfigure}

The energy spectral density, averaged power, and range are found to be the most important features in the mortality prediction based on the heart rate signal (Figure~\ref{fig:Predictor_importance}). As described above, the averaged power is one of the most important features since it is placed as the root of the decision tree. However, the energy spectral density gained the highest score of importance in comparison to the other features. Hence, the nodes corresponding to the energy spectral density feature have higher amount of $GDI$ compared to their children. As a matter of fact, this is a sign of high $GDI$ gap between these nodes and their children.

The energy spectral density provides basic information about the power variation in frequency components comprising the original signal within a finite interval. Since the power spectral density employs Fourier transform to decompose original signals into a spectrum of frequencies, it can reflect the parasympathetic and sympathetic activities which are highly correlated to the fluctuation of frequency components of heart signals. It has been reported~\cite{hasegawa_assessment_2015} that the high-frequency component reflects parasympathetic nervous activity, while the ratio of low-frequency over the high-frequency components reflects sympathetic nervous activity. Hence, a combination of frequency-domain (e.g. energy spectral density) and time-domain signal analysis (such as skewness) enables us to separate CCU patients who survive or pass away.

\section{Conclusion and Future Work}
  \label{sec:conclusion}
Early hospital risk of mortality prediction in CCU units is critical due to the need for quick and accurate medical decisions. This paper proposes a new signal-based model for early mortality prediction, leveraging the benefits of statistical and signal-based features. Our method is a clinical decision support system which focuses on using only the heart rate signal instead of other health variables such physical state or presence of chronic diseases. Since such variables require laboratory test results which could delay the decision-making time or may not be available at the time of admission, our proposed method may give faster feedback to healthcare professionals working in CCUs. We demonstrate the capability of using statistical and signal-based features, especially the energy-based features of heart rate signals, to distinguish between patients who survive or pass away in the CCU. Among the interpretable classifiers, the decision tree achieved the highest accuracy, allowing for both accurate and explainable outcomes.
 
In our future work, we plan to apply our proposed method over other intensive care units, incorporating multiple vital signals along with the heart rate signal as a means to better understand the cause of mortality. The study also can be extended to develop a framework using sensors, laboratory data, and information cached from intensivists and nurses' reports using  knowledge graph~\cite{shekarpour_rquery:_2017} and text mining~\cite{allahyari_brief_2017}. Another direction is to explore the effect of computing features from vital signals with different length of windows and using dynamic feature selections~\cite{zabihimayvan_soft_2017}\cite{hamidzadeh_detection_2018}. Finally, we plan on creating a real-time mortality prediction system based on the variability of physiological signals~\cite{kaffashi_variability_2007} that can predict patient outcomes for early intervention.

\section*{Supplementary Material}
  \label{sec:supplementary}

 The source code is available at: https://github.com/RezaSadeghiWSU/Early-Hospital-Mortality-Prediction-using-Vital-Signals

\section*{Acknowledgments}
  \label{sec:Acknowledgments}
This paper is based on work supported by the National Institutes of Health (NIH) under Grant no. $K01LM012439$. Any opinions, findings, and conclusions or recommendations expressed in this material are those of the author(s) and do not necessarily reflect the views of the NIH.

\bibliographystyle{elsarticle-num}

\bibliography{reference}





\end{document}